\documentclass[letterpaper, 10 pt, conference]{ieeeconf}  

\IEEEoverridecommandlockouts                              
\makeatletter
\def\bstctlcite{\@ifnextchar[{\@bstctlcite}{\@bstctlcite[@auxout]}}
\def\@bstctlcite[#1]#2{\@bsphack
  \@for\@citeb:=#2\do{%
    \edef\@citeb{\expandafter\@firstofone\@citeb}%
    \if@filesw\immediate\write\csname #1\endcsname{\string\citation{\@citeb}}\fi}%
  \@esphack}
\makeatother

\usepackage[vlined, ruled, linesnumbered, commentsnumbered]{algorithm2e}
\usepackage{amsmath}
\usepackage{cite}
\usepackage{color}
\usepackage{xcolor}
\definecolor{chred}{rgb}{0.8,0,0}
\definecolor{chgrey}{rgb}{0.5,0.5,0.5}
\usepackage{graphicx}
\usepackage{caption}
\usepackage{subcaption}
\usepackage{dblfloatfix}
\usepackage{eqlist}
\usepackage{txfonts}
\usepackage{url}
\usepackage{footmisc}
\usepackage{booktabs}
\usepackage{balance}

\usepackage{multirow}
\usepackage[para,online,flushleft]{threeparttable}
\title{\LARGE \bf Dual-arm Assembly Planning Considering Gravitational Constraints}

\author{Ryota Moriyama$^{1}$, Weiwei Wan$^{1}$$^{2}$$^{*}$, and Kensuke Harada$^{1}$$^{2}$%
\thanks{$^{1}${Graduate School of Engineering Science, Osaka University, Japan.}
$^{2}${National Inst. of AIST, Japan.} *{Correspondent author: Weiwei Wan:
}{\tt\small wan@sys.es.osaka-u.ac.jp}}
}

\begin{document}
\maketitle
\thispagestyle{empty}
\pagestyle{empty}

\begin{abstract}
Planning dual-arm assembly of more than three objects is a challenging Task
and Motion Planning (TAMP) problem. The assembly planner shall consider not
only the pose constraints of objects and robots, but also the
gravitational constraints that may break the finished part.
This paper proposes a planner to plan the dual-arm assembly of more
than three objects. It automatically generates the grasp configurations
and assembly poses, and simultaneously
searches and backtracks the grasp space and assembly space to accelerate the
motion planning of robot arms.
Meanwhile, the proposed method considers gravitational constraints during robot motion planning
to avoid breaking the finished part. In the experiments and analysis section, the
time cost of each process and the influence of different parameters used in the proposed planner are compared
and analyzed. The optimal values are used to perform real-world executions of various robotic assembly tasks.
The planner is proved to be robust and efficient through the experiments.
\end{abstract}


\section{Introduction} \label{introduction}
Planning dual-arm assembly of mutiple objects is a challenging problem.
During the assembly, one of the robot arms holds the finished part, and the other
arm assembles the next part to it. To automatically plan the assembly motion,
one needs to consider the grasp configurations, the assembly positions and orientations,
the constraints from the start and goal poses of the object parts,
the kinematic constraints of the robots, as well as the gravitational constraints
of the finished part. While these problems were previously studied independently
or partially, they are not analyzed and developed as an integral planner for dual-arm multi-object assembly. 

Under this background, this paper develops an integral planner to plan the dual-arm assembly of more
than three objects. The planner automatically generates the grasp configurations for each object and
the assembly positions and orientations in the work space. The grasp configurations and assembly positions/orientations
are sampled from a grasp space and an assembly space. The proposed planner simultaneously
searches and backtracks the grasp space and assembly space to accelerate the
motion planning of robot arms.

\begin{figure}[!htbp]
    \centering
    \includegraphics[width=\columnwidth]{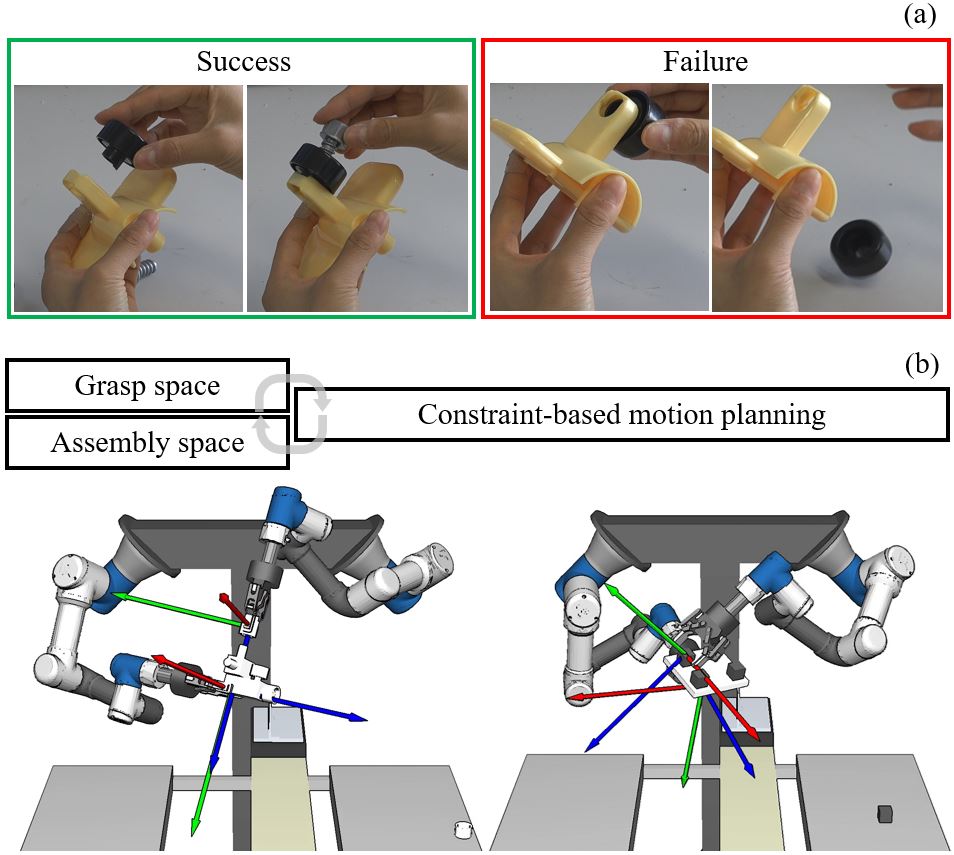}
    \caption{(a) In the left part, the holding
    hand keeps the yellow object tilted to support the wheel. The bolt is thus
    successfully assembled. In the right part, the wheel drops down due to pose changes and gravity.
    (b) Upper part: The workflow of the proposed planner; Lower part: Two exemplary keyposes
    from the planned motion.}
    \label{fig: teaser}
\end{figure}

Especially, the proposed planner is able to deal with the planning of more than three objects
by considering gravitational constraints. Gravity may break the finished part when the 
holding hand changes its pose without considering the gravitational constraint.
Fig.\ref{fig: teaser}(a) shows an example. The left part shows a successful assembly motion where the holding
hand kept the yellow object tilted to support the wheel. The left hand could release and continue to
assemble the bolt. In contrast, the right part shows a failure motion. In this case, the left hand moved,
the wheel dropped down, and the finished part was broken.
The planner proposed in this paper aims to avoid the failure happened in the second case by considering gravitational constraints.

Fig.\ref{fig: teaser}(b) shows an overview of the proposed planner. 
It loops through the grasp space and assembly space (the contents
in the left two boxes in the upper part of Fig.\ref{fig: teaser}(b)) to search and backtrack
the grasp configurations and assembly poses of dual-arm assembly. Especially
during the motion planning (the right box in the upper part of Fig.\ref{fig: teaser}(b)),
the planner considers both collision and
gravitational constraints to not sample problematic nodes and ensure safe motion.

In the experiments and analysis section, the
time cost of each process and the influence of different parameters used in the proposed planner are compared
and analyzed. The optimal values are used to perform real-world executions of various robotic assembly tasks.
The planner is proved to be robust and efficient through the experiments.

\section{Related Work}\label{related_work}

The related studies of this work include: classic assembly planning, robotic task and motion planning, and dual-arm manipulation.

\subsection{Classic assembly planning}\label{assemblyplanning}
The seminal studies in classic assembly planning focused on the high-level geometric reasoning
of CAD models \cite{de1987simplified}\cite{de1990and}\cite{wilson1994geometric}.
The goal was to generate the order in which parts come together.
Similar following studies include \cite{sanderson1999assemblability}\cite{knepper2013ikeabot}\cite{jimenez2013survey}.
These early assembly planning studies used mostly geometric constraints.

More recent planners tend to use a mixed model of constraints.
The seminal studies of classic assembly planning 
were extended by considering gravitational
constraints and the mutual support between objects in the 1990s \cite{mattikalli1995gravitational},
and lots of studies were inspired by the extension. For example,
Dobashi et al.\cite{dobashi2014robust} considered both geometric and grasp constraints to assemble wooden blocks.
Dogar et al.\cite{dogar2015multi} used both geometric and grasp constraints to assemble chairs.
Suarez-Ruiz et al.\cite{suarez2018can} presented a similar chair-assembly task, where force control was further included to finish the precise insertion.
MacEvoy et al.\cite{mcevoy2014assembly} considered the stability and geometric constraints to plan assembly sequences
of truss structures.
A survey that summarized the work that considered various constraints before 2015 is
available in \cite{ghandi2015review}. In our recent work\cite{wan2018assembly}, 
a mixed constraint of stability, graspability, and assemblability
was used to plan the robot motion to assemble Soma cubes.

This paper assumes the order of assembly is given. It does not re-implement
a classic assembly planner that uses various constraints to optimize assembly orders.
Instead, the constraints are considered during the planning of grasp configurations,
assembly poses, and robot motion, which is more like the task and motion planning that follows.

\subsection{Task and motion planning} 
Task and motion planning (TAMP) plans the robotic motion to manipulate parts
considering geometric constraints and task orders \cite{kaelbling2013integrated}. It includes
but is not limited to assembly. 

Modern TAMP planners use a symbolic reasoner to find task-level jobs, and iteratively runs
motion planning to find the robot motion for each job. For example,
Wolfe et al. \cite{wolfe2010combined} used HTN (Hierarchical Task Networks) to divide and conquer sub-planning problems.
Srivastava et al. \cite{srivastava2014combined} interweaved PDDL (Planning Domain Description Language) and RRT (Rapidly-exploring Random Trees) to plan
a motion to arrange a cluttered table. Zhang et al. \cite{zhang2016co} used multi-level
optimization, incorporating task, action, and motion planning, to generate a combined motion plan for a mobile manipulator.
Recently, Lagriffoul et al. \cite{lagriffoul2018platform} proposed a benchmark for
TAMP planners. The benchmark includes not only mobile pick-and-place tasks but also assembling parts at a stationary position.

The planner developed in this paper is one kind of TAMP planner. 
Compared to the other TAMP planners, our main contribution is planning the pick-and-assembly motion
of multi-objects using dual-arm robots.

\subsection{Dual-arm manipulation}
Dual-arm manipulation and dual-arm manipulators are popular topics in robotics and are
widely studied\cite{siciliano2012advanced}\cite{park2016dual}. Most of the dual-arm manipulation planning studies focus
on sequential manipulation\cite{saut2010planning}\cite{wan2018regrasp}\cite{chen2018manipulation} and
coordinated dual-arm motion planning\cite{cohen2012search}\cite{ramirez2017human}.
Dual-arm assembly is less studied. The most recent dual-arm assembly work we could find is \cite{stavridis2018bimanual},
which used dual-arm robots to assemble two parts considering motion and force capabilities.

This paper develops a dual-arm assembly planner. The difference from previous dual-arm manipulation studies is
the consideration of multiple parts and gravitational constraints.
An integral and practical planner for dual-arm multi-object assembly is implemented. 

\section{Overview of the Method}\label{method}

\begin{figure}[!htbp]
    \centering
    \includegraphics[width=\columnwidth]{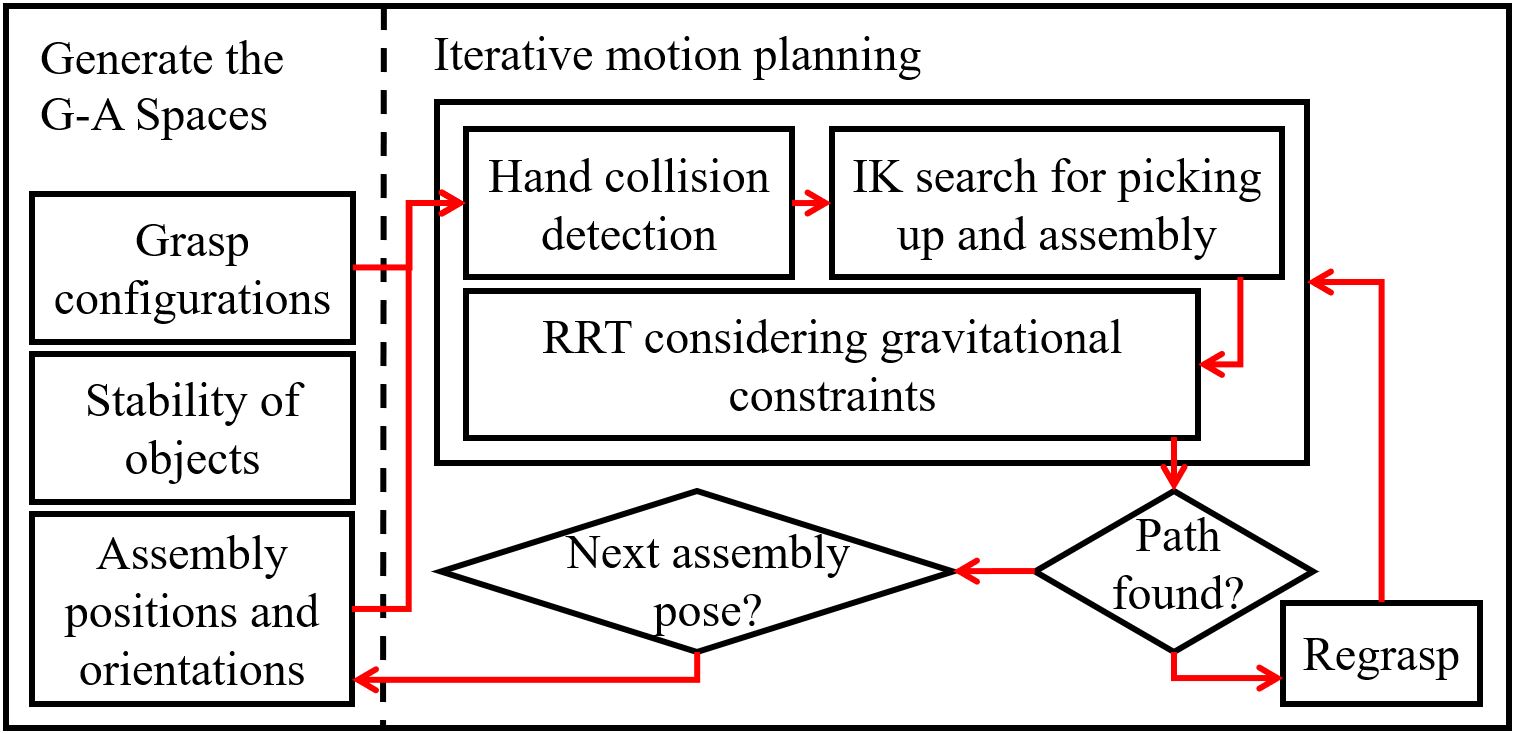}
    \caption{Detailed workflow of the proposed method. It iterates through the grasp space and assembly spaces (G-A spaces)
    to get a sequence of start and goal configurations for RRT. In case no path is found, the 
    method tries the next assembly positions and orientations, or refers to regrasp to reorient the initial poses of the parts.}
    \label{fig: flow}
\end{figure}

This section gives a general explanation of the proposed method. The rough idea is already mentioned in
the upper part of Fig.\ref{fig: teaser}(b), which iterates through the grasp space and assembly space to
get the start and goal configurations for sampling-based motion planning. The detailed workflow is shown in Fig.\ref{fig: flow}.

First, the method generates the grasp and assembly spaces (G-A spaces) by synthesizing grasp configurations and assembly poses,
as is shown in the left part of Fig.\ref{fig: flow}. The details of the synthesizing will be discussed in Section IV.
The stability of the parts, namely the stable placements of the parts on tables, is also computed to allow the robot
to reorient them using regrasp planning\cite{wan2018regrasp}.

The right part of Fig.\ref{fig: flow}, namely the iterative planning part will
search the G-A spaces to get a sequence of start and goal configurations
for RRT motion planning. If the sequence of start and goal configurations could be found,
the planner will call gravity-constrained RRT to plan robot motion between each adjacent start and goal
in the sequence. If the sequence is not available, the planner will iterate to the next assembly pose,
and re-search. The method reports a successful plan if all robot motion between the adjacent start and goal
configurations in a sequence could be found. 
The details of iteration and gravity-constrained RRT will
be discussed in Section V. Especially, two different algorithms were proposed to select the start and goal configurations.
These two algorithms are compared in detail in the experiments and analysis section.

\section{The Grasp and Assembly Spaces (G-A Spaces)}\label{spaces}

\begin{figure}[!htbp]
    \centering
    \includegraphics[width=\columnwidth]{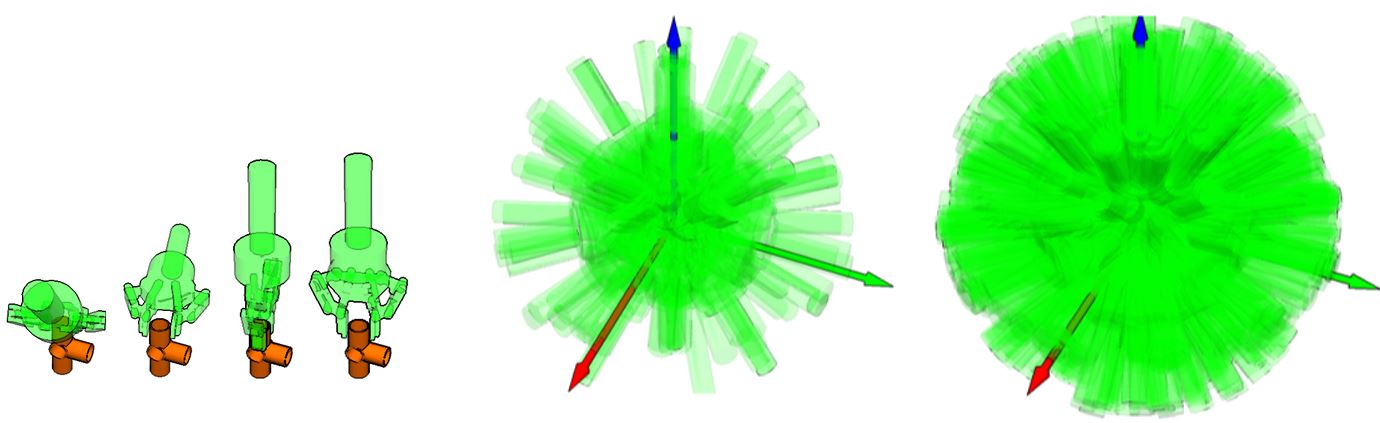}
    \caption{Some grasp configurations sampled from the grasp space. The left figure shows 
    several independent grasp configurations.
    The middle and right figures show two final results.
    The middle one has a smaller number of synthesized grasp configurations than the right one.
    The number of grasps could be controlled by tuning the parameters provided by the grasp synthesis method.}
    \label{fig: grasps}
\end{figure}

The grasp space is defined as a space of hand poses that can hold an object in force closure.
We use a grasp synthesis method proposed in \cite{wan2016developing} to compute the grasp space of an object.
The method is able to plan precise grasp poses for suction cups and parallel grippers.
The basic idea is to find planar facets, sample the facets, and find the candidate samples
for attaching suction cups or gripper finger pads. The method provides several tunable parameters to
control the density of the synthesized grasp configurations. By using this method,
we can automatically compute a lot of candidate grasps using the CAD model of an object.
Fig.\ref{fig: grasps} shows some grasp configurations synthesized by the method. They 
are discretized samples from the grasp space.

\begin{figure}[!htbp]
    \centering
    \includegraphics[width=\columnwidth]{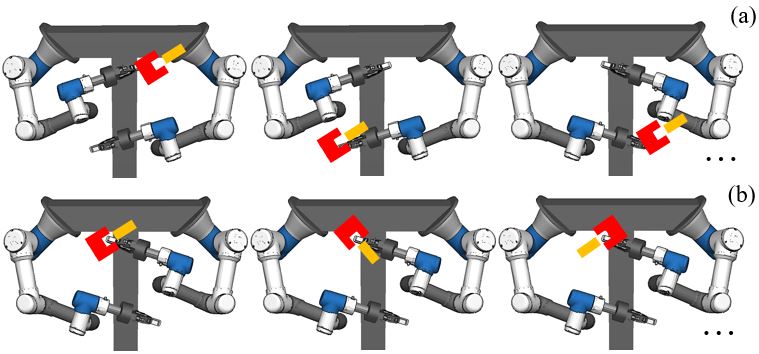}
    \caption{Assembly positions and orientations. The robot may assemble the yellow peg 
    and the red slot at lots of different positions shown in Fig.\ref{fig: assembly}(a), 
    as well as lots of different orientations shown in Fig.\ref{fig: assembly}(b).}
    \label{fig: assembly}
\end{figure}

\begin{figure}[!htbp]
    \centering
    \includegraphics[width=\columnwidth]{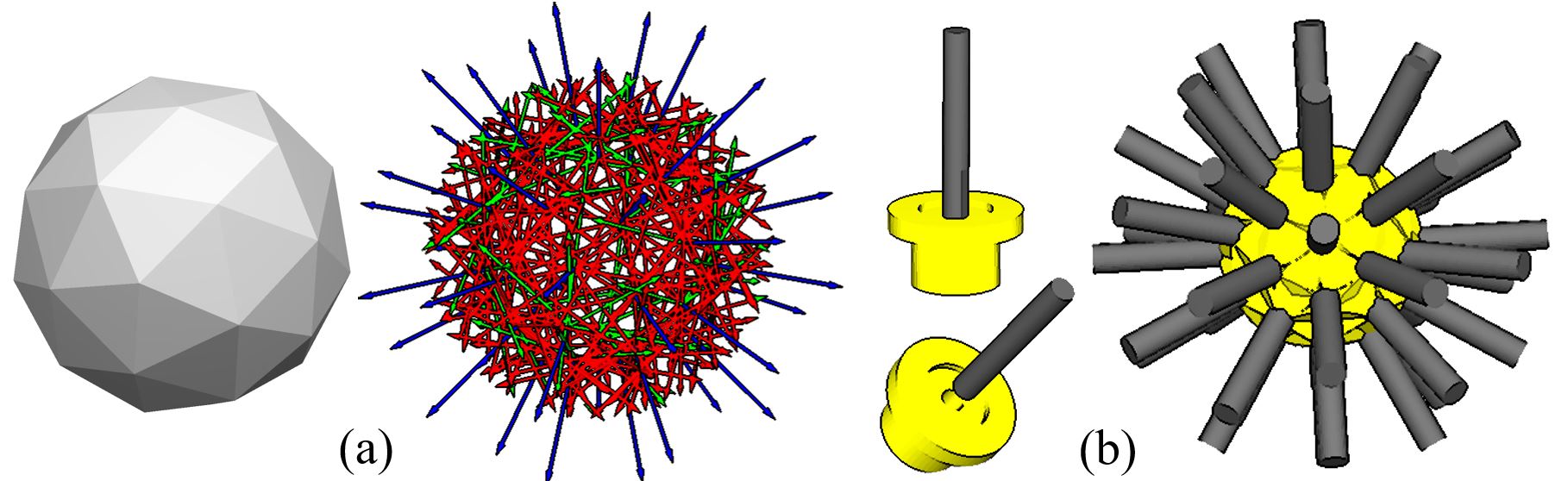}
    \caption{(a) Left: A level-1 icosphere. Right: The assembly orientations obtained by defining rotation frames
    using the vectors pointing to the vertices of the icosphere. (b) An example where the shaft and bearing in the left part is sampled to be
    assembled at the many candidate orientations shown in the right.}
    \label{fig: icosphere}
\end{figure}

The assembly space is defined as a space of positions and orientations for the dual-arm
robot to assemble two objects. Fig.\ref{fig: assembly} shows an example. The dual-arm robot may
assemble a yellow peg and a red slot at lots of different positions shown in Fig.\ref{fig: assembly}(a), as well as lots of different 
orientations shown in Fig.\ref{fig: assembly}(b). These different positions and orientations are discretized samples from
the assembly space. To plan these discrete assembly positions and orientations, random
sampling together with icospheres are used. First, we discretize the candidate assembly region by
defining an area in front of the dual-arm robot, and sample the assembly region to get several
assembly positions. Then, at each sampled position we sample the assembly orientations
by using icospheres (see the left part of Fig.\ref{fig: icosphere}(a)). The vector pointing to each vertex of 
an icosphere is used as the $z$ axis of an assembly orientation (blue vectors in the right part of Fig.\ref{fig: icosphere}(a)).
The $x$ and $y$ are sampled around the $z$ axis to finalize the rotation frames (the red and green vectors in the right part of Fig.\ref{fig: icosphere}(a)).
Different levels of icospheres are used and analyzed in the experiments and
analysis section to show their influence on the performance of the proposed planner.
Fig.\ref{fig: icosphere}(b) shows an example where a shaft and bearing in the left part is sampled to be
assembled at the many candidate orientations shown in the right. These candidate orientations
are obtained from the rotation frames shown in Fig.\ref{fig: icosphere}(a).

\section{Motion Planning by Interacting with the G-A Spaces}\label{planning}

\subsection{Assembling the first two objects}\label{planning-a}
\begin{figure}[!htbp]
    \centering
    \includegraphics[width=\columnwidth]{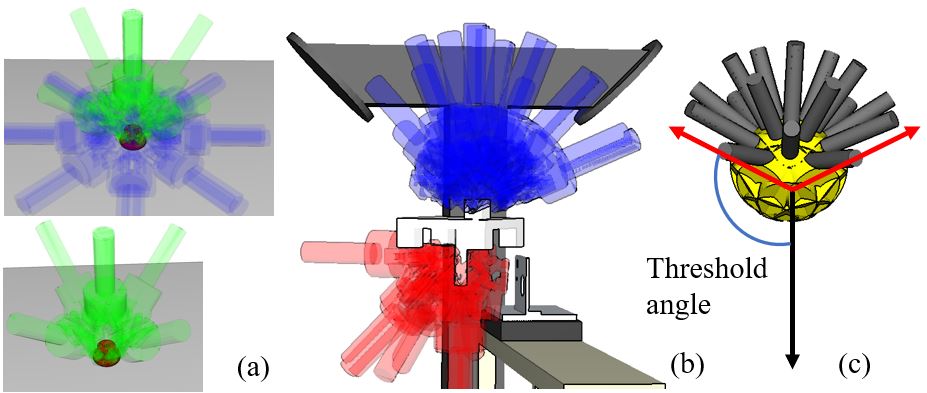}
    \caption{(a) Selecting the grasp configurations to pick up an object
    from its initial pose. The green hands are the selected grasps. (b) Selecting an assembly pose and finding the collision-free grasp configurations for the two hands.}
    \label{fig: selgrasp}
\end{figure}

After sampling the G-A spaces and obtaining the discretized grasp configurations and
assembly poses, the proposed method starts iterating through them to get a sequence of key
robot poses and plan the robotic assembly motion. The right part in Fig.\ref{fig: flow}
roughly shows the workflow of the iteration. First, the ``Hand collision detection'' frame box will select
some candidate grasp configurations and an assembly pose from the discretized G-A space.
All the objects, the surrounding obstacles, and the kinematics of the robot are considered during the selection.
The candidate collision-free grasp configurations for objects both at
the initial pose and the assembly pose will be obtained after the selection.
An exemplary result is shown in Fig.\ref{fig: selgrasp}.
In the left part, a bearing is at its initial pose on a table.
The ``Hand collision detection'' frame box selects the green hands as the candidates to pick up
the bearing. The blue hands are the collided hands and are removed.
In the right part, the ``Hand collision detection'' frame box selects one assembly pose and finds
the collision-free grasp configurations to assemble the parts. The red hands are
the candidates grasp configurations for the right hand. The blue hands are the candidate grasp configurations for the left hand.
The proposed method will solve the IK for the candidate grasp configurations and use
the found robot poses as start and goals for RRT motion planning. They are carried out in
the ``IK search for picking up and assembly'' frame box, and the ``RRT considering gravitational constraints'' frame box.

\begin{figure}[!htbp]
    \centering
    \includegraphics[width=\columnwidth]{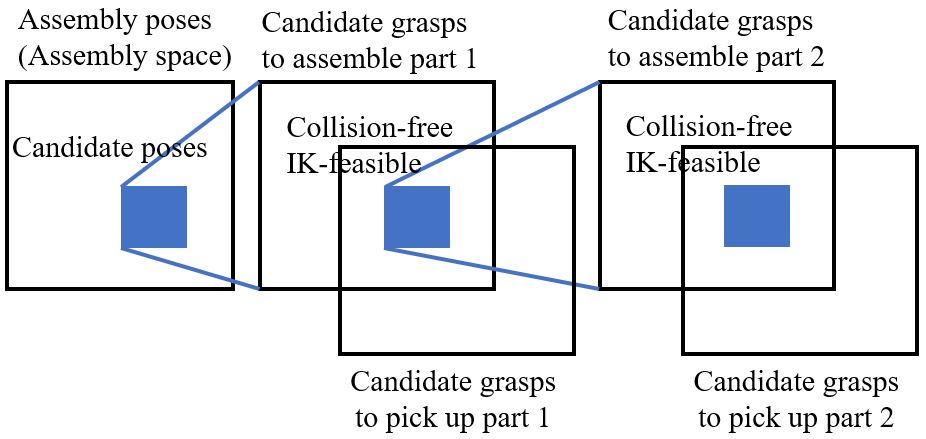}
    \caption{The first way to select the grasps for assembly.}
    \label{fig: select0}
\end{figure}

\begin{figure}[!htbp]
    \centering
    \includegraphics[width=\columnwidth]{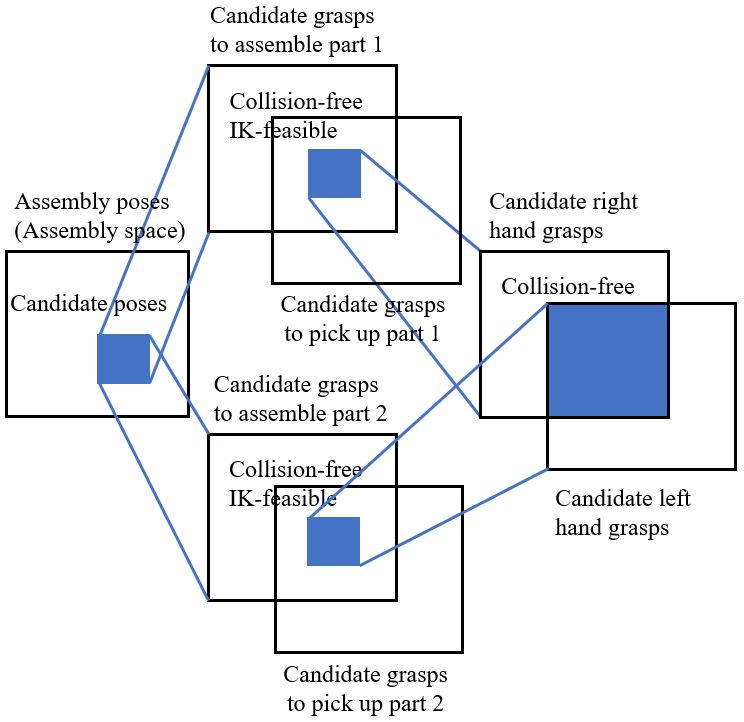}
    \caption{The second way to select the grasps for assembly.}
    \label{fig: select1}
\end{figure}

The order to perform the selection is crucial to the time cost.
There are two ways to implement the order. 
They are shown in Fig.\ref{fig: select0} and Fig.\ref{fig: select1} respectively.
The two parts in the assembly are named part 1 and part 2.
In the first case, the algorithm works as follows.
\begin{itemize}
\item[(1)] The algorithm chooses a candidate assembly pose from the assembly space.
\item[(2)] For part 1, the algorithm computes the ``candidate grasps to assemble part 1'' and the
``candidate grasps to pick up part 1'' from its initial pose.
\item[(3)] The algorithm computes the intersection of the ``candidate grasps to assemble part 1'' and the
``candidate grasps to pick up part 1', and finds a set of collision-free and IK-feasible grasps from the intersection. 
\item[(4)] For each grasp in the set, the algorithm repeats (1)-(3) for part 2, avoiding not only
the collision with the parts, but also the collision with the candidate grasp of part 1. The algorithm stops
if a feasible grasp for part 2 is found.
\end{itemize}

The workflow of the second algorithm is as follows.
\begin{itemize}
\item[(1-3)] The same as the first algorithm.
\item[(4)] The algorithm performs (3) and (4) for part 2.
\item[(5)] The algorithm merges the CD-free and IK-feasible grasps of part 1 and part 2 by pairing them.
The algorithm returns the first pair for future use.
\end{itemize}

The first algorithm repeatedly performs collision detection for each of the selected grasps for part 1.
The advantages of the algorithm is it may quickly find a pair of candidate grasps for the two parts
if a solution exists and the random search is lucky. In contrast, the second algorithm computes the collision-free and IK-feasible grasps for both part 1 and part 2
in the second, and uses a pairing step to merge them. It is faster to find all pairs of candidate grasps, but is slower to get a single one
since all grasps for the two parts have to be examined before getting the results. Both the two algorithms are implemented in our planner.
Their performance is compared in the experimental section.

In the two-object assembly, we do not really need to consider gravitational constraints during motion planning
since all objects are held firmly by the two hands. 
We only need to consider gravity when choosing the candidate assembly poses (the first step of the two algorithms)
-- Part 2 must be supported by part 1 to avoid dropping down.
This constraint is defined as an angle between the assembly direction and gravitational direction. When the angle is smaller
than a value, the assembly is considered to be unstable. The assembly poses of the peg and bearing,
after considering the gravitational constraints, are shown in Fig.\ref{fig: selgrasp}(c). The peg never points downward.

\subsection{Assembling the remaining parts}
\begin{figure}[!htbp]
    \centering
    \includegraphics[width=.9\columnwidth]{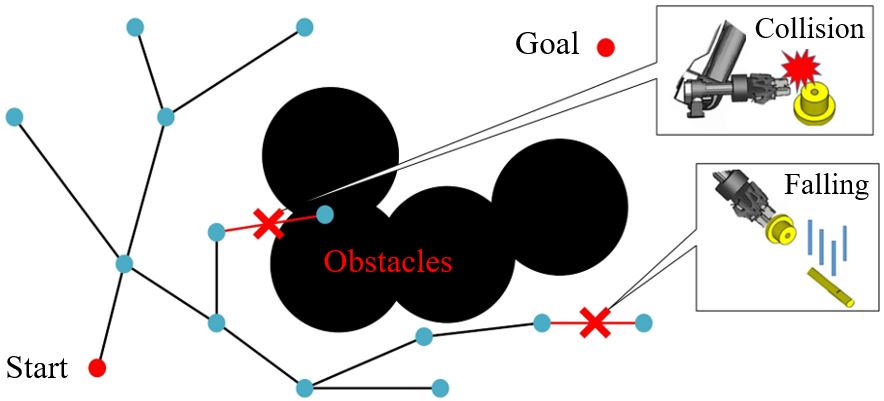}
    \caption{Gravity-based motion planning. Both collision and the gravitational constraints are
    examined when a new node is sampled during the motion exploration.}
    \label{fig: rrt}
\end{figure}

When assembling the third part, part 1 and part 2 are assembled together. They become
a finished part. The goal of assembly is to assemble part 3 to the finished part (namely part 12). 
The selection of candidate grasps and the motion planning for the arm holding part 3 is the same as the two-object assembly. 
The motion planning for the arm holding part 12 needs to be carefully designed considering the gravitational constraints.
Our implementation is the gravity-based motion planning illustrated in Fig.\ref{fig: rrt}. It is a variation of RRT motion planning. When sampling a new
node, the gravity-based planner not only check the collision of the node and the edge with the configuration obstacles, but also the stability of the finished part.
The gravity-based planner ensures that part 2 will be supported by part 1 to avoid falling down.

The gravitational constraint is also implemented as the angles between constraint vectors.
Assume the assembly direction of part 12 is denoted by $v_{a12}$. The gravitational constraint is to
require the angle between $v_{a12}$ and gravitational direction to be larger than a threshold. If the angle is too small, the
assembly is considered to be unstable. For the 4th, 5th, etc., objects, the gravitational constraint is met when
all angles between the assembly directions and gravitational direction are larger than the given threshold.

\section{Experiments and Analysis}

\begin{figure}[!htbp]
    \centering
    \includegraphics[width=.95\columnwidth]{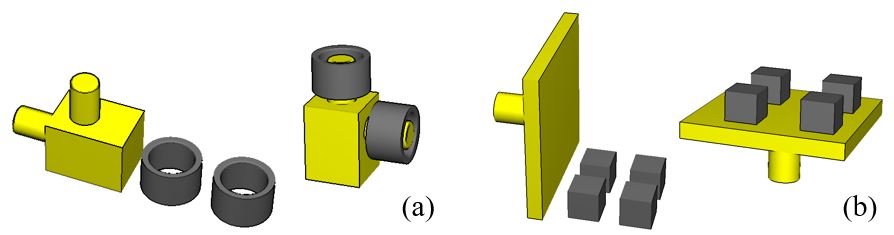}
    \caption{Two sets of assembly tasks used in the experiments.
    The first task is to assemble one base and two rings shown in the left part of (a) to
    the right state. The second task is to place four blocks onto a plate.}
    \label{fig: sets}
\end{figure}

We developed both simulation and real-world experiments to examine the proposed planner.
Two sets of assembly tasks, shown in Fig.\ref{fig: sets} were used for analysis.
The first set was to put two rings on a branched base. The rings were quite loose.
A robot must not let any branch face downward during assembly. The second set was to
place four blocks onto a plate. A robot had to keep the plate facing upward to avoid losing 
finished blocks. The computational platform used in the experiments was a PC with Intel Core it-6500 CPU and 8.00GB memory.
The programming language was Python 2.7. The software platform was Pyhiro\footnote{https://github.com/wanweiwei07/pyhiro}.

\begin{figure*}[!htbp]
    \centering
    \includegraphics[width=.95\textwidth]{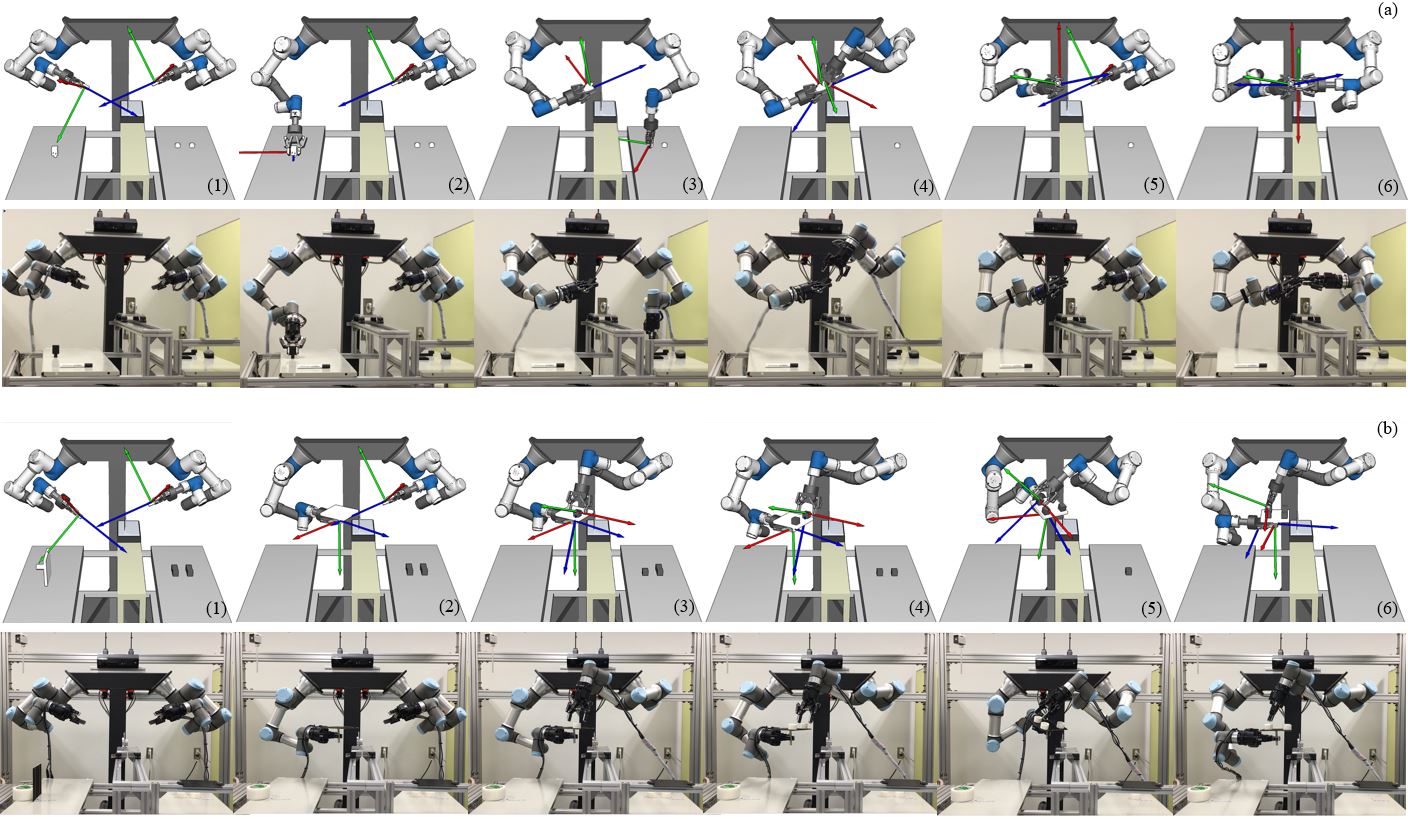}
    \caption{Two sets of assembly tasks used in the experiments.
    The first task is to assemble one base and two rings shown in the left part of (a) to
    the right state. The second task is to place four blocks onto a plate.}
    \label{fig: results}
\end{figure*}

The results of simulation and real-world execution are shown in Fig.\ref{fig: results}.
For the first assembly set, the robot chose to assemble the first ring from an upward direction, 
as is shown in Fig.\ref{fig: results}(a-3,4). Before assembling the third ring, the robot rotated the finished part in Fig.\ref{fig: results}(a-5)
while successfully avoided losing the first ring. The gravity-constrained planner was acting an important role in 
producing the motion. For the second assembly set, The robot was maintaining the upward pose of the plate
in Fig.\ref{fig: results}(b-3,4,5,6). At the same time, it was also computing the optimal pose for assembly motion. In Fig.\ref{fig: results}(b-3,4,6),
robot was holding the plate horizontally. In Fig.\ref{fig: results}(b-5), the robot tilted the plate a bit to let its left arm easily access
the goal dropping position. The gravity-constrained planner optimized the assembly poses automatically.

The parameters used in the experiments are shown in Table \ref{Tab:parameters}.
The number of sampled grasp configurations for the right and left arms was 200 and 234 respectively.
The number of sampled assembly positions was 4. The number of sampled assembly orientations was 252 (icosphere level 2).
The angle threshold for the first assembly set was 90$^\circ$. It required the angle between a branch
and the gravitational direction to be larger than 90$^\circ$. The angle threshold for the second assembly set was
175$^\circ$. It required the angle between the upward direction of the plate and the gravitational direction to 
be larger than 175$^\circ$. The second selection algorithm mentioned in Section \ref{planning-a} was used.

\begin{table}[!htbp]
\tiny
\centering
\caption{The parameters used in the experiments}
\label{Tab:parameters}
\resizebox{\columnwidth}{!}{%
\begin{threeparttable}
\begin{tabular}{lll}
\toprule
G space & \# of sampled grasps (right$\cdot$left) & 200 $\cdot$ 234
\\
\multirow{ 2}{*}{A space} & \# of sampled assembly positions & 5
\\
& \# of sampled assembly orientations & 14913
\\
\midrule
Grav. cstr. & Set 1: 90$^\circ$ & Set 2: 175$^\circ$
\\
Sel. ord. & Order 2 &\\
\bottomrule
\end{tabular}
\begin{tablenotes}
\# of sampled grasps (right$\cdot$left): \# of right hand $\cdot$ \# of left hand; Grav. cstr. - Gravity constraint;
Sel. ord. - Selection order.
\end{tablenotes}
\end{threeparttable}}
\end{table}

The computational costs to plan the assembly motion for the two sets are shown in Table \ref{Tab:real_world_time}.
The time is the mean of ten times of simulation. The problems could be solved safely with less than 1 minute,
using pure Python.

\begin{table}[!htbp]
\tiny
\centering
\caption{Time costs of the two assembly sets}
\label{Tab:real_world_time}
\resizebox{\columnwidth}{!}{%
\begin{threeparttable}
\begin{tabular}{lcccc}
\toprule
 & Set 1 & Set 2 
\\
\midrule
First assembly pose and IK & 34.95 $s$ & 12.34 $s$
\\
The remaining assembly poses and IK & 18.23 $s$ & 7.84 $s$
\\
First gravity-constrained RRT & 42.74 $s$ & 21.73 $s$
\\
The remaining gravity-constrained RRT & 26.58 $s$ & 17.69 $s$
\\
\bottomrule
\end{tabular}
\begin{tablenotes}
The time costs in the table are the mean of ten simulation.
\end{tablenotes}
\end{threeparttable}}
\end{table}

Besides the chosen parameters shown in Table \ref{Tab:parameters}, various
other candidates were tested.
The computational costs under the other settings and 
the detailed costs of each process in the planner are shown as follows. First, 
for the two selection orders mentioned in Section \ref{planning-a}. The costs are shown in
Table \ref{Tab:order}. The second order was more stable when the random search was not lucky.
Also, the second order produced more flexible right-left combinations than the first order.
Multiple right grasps were paired with multiple left grasps. In contrast, one right grasp
was paired with multiple left grasps in the first order.
Thus, the second selection algorithm was used.

\begin{table}[!htbp]
\tiny
\centering
\caption{Comparison of the two selection orders}
\label{Tab:order}
\resizebox{.85\columnwidth}{!}{%
\begin{threeparttable}
\begin{tabular}{lcccc}
\toprule
 & Set 1 & Set 2 
\\
\midrule
With IK-feasible grasps & 1.14 $s$ & 6.31 $s$
\\
IK-feasible at the 10th trial & 150.17 $s$ & 88.96 $s$
\\
Candidate pairs (right$\cdot$left) & 1$\cdot$n & m$\cdot$n
\\
\bottomrule
\end{tabular}
\begin{tablenotes}
The time costs in the table are the mean of ten simulation.
\end{tablenotes}
\end{threeparttable}}
\end{table}

The time costs with respect to different G-A space samples are shown in Table \ref{Tab:ga}.
The left column is the number of sampled assembly positions. The upper row is the number
of sampled grasps. As the number of grasps increased, the time costs got much larger.
On the other hand, the number of assembly positions did not significantly increase the computational load.
For this reason, we used 4 in the experiments.

\begin{table}[!htbp]
\tiny
\centering
\caption{Comparison of the two selection orders}
\label{Tab:ga}
\resizebox{\columnwidth}{!}{%
\begin{threeparttable}
\begin{tabular}{cccccc}
\toprule
\multirow{ 2}{*}{} & \multicolumn{5}{c}{\# of sampled grasps (right $\cdot$ left)}
\\ \cmidrule{2-6}
\# posn. & 200 $\cdot$ 234 & 256 $\cdot$ 348 & 320 $\cdot$ 456 & 420 $\cdot$ 614 & 596 $\cdot$ 772 
\\
\midrule
3 & NA & 91 $s$ & 124 $s$ & 144 $s$ & 212 $s$
\\
4 & 90 $s$ & 94 $s$ & 130 $s$ & 149 $s$ & 227 $s$
\\
5 & 92 $s$ & 101 $s$ & 133 $s$ & 152 $s$ & 249 $s$
\\
\bottomrule
\end{tabular}
\begin{tablenotes}
The time costs in the table are the mean of ten simulation.
\end{tablenotes}
\end{threeparttable}}
\end{table}

The detailed costs and discovery rates for different numbers of assembly orientation are shown in
Table \ref{Tab:orien}. Here, different levels of icospheres are used to generate the assembly poses.
The correspondent illustrations of the different levels are under the table. The results showed
that the success rate increased as the number of orientation increased. Meanwhile, there was
no significant change in the time cost before
finding feasible assembly motion by random search. Thus, we used the highest number of orientation, 252, in the results
shown in Fig.\ref{fig: results}.

\begin{table}[!htbp]
\tiny
\centering
\caption{Comparison of different \# of assembly orientation}
\label{Tab:orien}
\resizebox{\columnwidth}{!}{%
\begin{threeparttable}
\begin{tabular}{lcccc}
\toprule
Icosphere level & \# orien. & $t$ full & Discov. rate & $t$ discov.
\\
\midrule
Tetrahedron & 24 & 1500 $s$ & 16.7 \% & 390 $s$
\\
Hexahedron & 48 & 3203 $s$ & 14.5 \% & 419 $s$
\\
Octahedron & 36 & 1924 & 13.9 \% & 334 $s$
\\
Icosahedron & 72 & 4447 & 18.3 \% & 370 $s$
\\
Icosphere Lv1 & 120 & 7454 & 16.7 \% & 341 $s$
\\
Icosphere Lv2 & 252 & 14913 & 19.4 \% & 304 $s$
\\
\bottomrule
\end{tabular}
\begin{tablenotes}
\# orien. - \# of assembly orientation; $t$ full - Time cost to explore all the orientation;
Discov. rate - success rate to find feasible assembly motion; $t$ discov. - The time cost before
finding feasible assembly motion by random search. 
The time costs and discovery rates in the table are the mean of ten simulation.
The different orientations used in Table \ref{Tab:orien} are as follows. From left to right: Tetrahedron,
Hexahedral, Icosahedron, Icosphere Lv1, Icosphere Lv2.
\begin{minipage}{.31\textwidth}
      \includegraphics[width=\columnwidth]{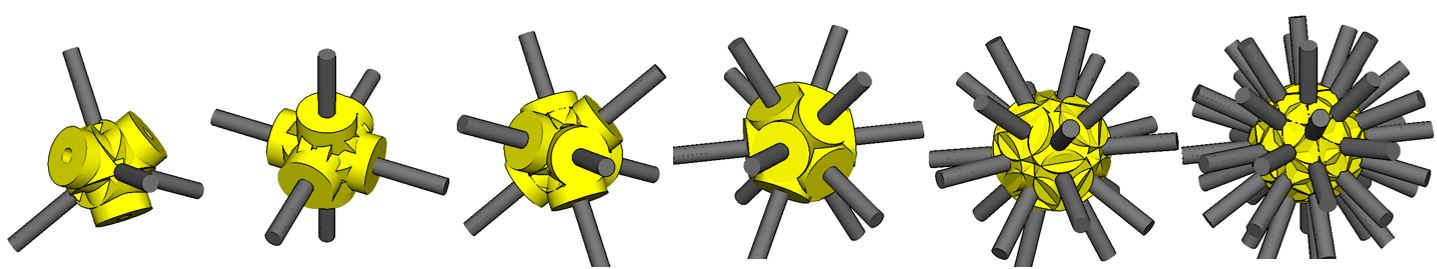}
\end{minipage}

\end{tablenotes}
\end{threeparttable}}
\end{table}

Finally, the detailed costs of each process in the planner with respect to different numbers of sampled grasps
are shown in Table \ref{Tab:process}. Here, the number of assembly positions was 4. The number of assembly
orientations was 252. Of all the process, the CD (collision detection) costed the most computational resources.
The IK costs increased significantly as the number of grasps increased. The RRT did not change much with more grasps.
Besides the presented results, readers may find some other examples in the attached video.

\begin{table}[!htbp]
\tiny
\centering
\caption{Comparison of the two selection orders}
\label{Tab:process}
\resizebox{\columnwidth}{!}{%
\begin{threeparttable}
\begin{tabular}{cccccc}
\toprule
\multirow{ 2}{*}{} & \multicolumn{5}{c}{\# of sampled grasps (right $\cdot$ left)}
\\ \cmidrule{2-6}
Process & 200 $\cdot$ 234 & 256 $\cdot$ 348 & 320 $\cdot$ 456 & 420 $\cdot$ 614 & 596 $\cdot$ 772 
\\
\midrule
CD1 & 20 $s$ & 22 $s$ & 34 $s$ & 41 $s$ & 63 $s$
\\
IK1 & 9 $s$ & 11 $s$ & 20 $s$ & 23 $s$ & 41 $s$
\\
RRT1 & 33 $s$ & 32 $s$ & 36 $s$ & 39 $s$ & 43 $s$
\\
CD2 & 8 & 10 $s$ & 16 $s$ & 20 $s$ & 36 $s$
\\
IK2 & 5 $s$ & 5 $s$ & 9 $s$ & 9 $s$ & 17 $s$
\\
RRT2 & 15 $s$ & 14 $s$ & 15 $s$ & 17 $s$ & 27 $s$
\\
\midrule
SUM & 90 $s$ & 94 $s$ & 130 $s$ & 149 $s$ & 227 $s$
\\
\bottomrule
\end{tabular}
\begin{tablenotes}
The time costs in the table are the mean of ten times of simulation.
\end{tablenotes}
\end{threeparttable}}
\end{table}

\section{Conclusions}\label{conclusion}

A dual-arm assembly planner to assemble multiple objects
considering gravitational constraints is developed. The planner selects grasp configurations and assembly poses
, as well as plans robot motion, considering gravity. It automatically picks
upward poses to assemble the second part, and rotates the finished part considering
gravity constraints to avoid dropping. Experiments and analysis
showed the proposed planner is able to optimize the poses and motion for safe assembly. The planner is proved to be robust and efficient for dual-arm assembly planning.


\bibliographystyle{IEEEtran}
\bibliography{papermoriyama}

\end{document}